%% file: main.tex
\documentclass[cameraready]{Interspeech}

\title{
On the Effect of Segmentation Width and Cluster Size on Speech Resynthesis\\
and Continuation in Generative Spoken Language Models
}

\author[affiliation={1}, orcid=0009-0004-8060-5583]{Shunsuke}{Kando}
\author[affiliation={1}, orcid=0009-0002-3454-0809]{Wataru}{Nakata}
\author[affiliation={2,1}, orcid=0000-0003-0520-7847]{Shinnosuke}{Takamichi}
\author[affiliation={1}, orcid=0000-0001-7271-0253]{Yusuke}{Miyao}

\address{
    $^1$ The University of Tokyo, Japan \\
    $^2$ Keio University, Japan
}

\email{skando@is.s.u-tokyo.ac.jp}


\usepackage{comment}
\usepackage{amsmath}
\usepackage{graphicx}
\usepackage{hyperref}
\usepackage{color}
\usepackage{booktabs}
\usepackage{makecell}
\usepackage{multirow}
\usepackage{amssymb}
\usepackage{pifont}
\usepackage{bm}
\usepackage{cite}

\begin{document}

\maketitle

\begin{abstract}
Generative Spoken Language Modeling (GSLM) enables text-free speech modeling by training language models (LMs) using discrete speech representations instead of textual transcription.
In this paper, we investigate the performance of GSLM on speech synthesis and continuation using discrete speech representations with varying bitrates.
We segment speech representations with fixed widths and train K-means models in multiple cluster sizes, resulting in various bitrate settings.
We demonstrate that intelligible and natural speech can be synthesized at lower bitrate settings than the baseline.
Furthermore, speech continuation quality remains stable at lower bitrates across multiple metrics, suggesting that the conventional GSLM setting may be redundant for effective speech generation.
Although LLM-based metrics show higher correlation with human subjective score than conventional metrics, it remains low, highlighting the need for more stable automatic evaluation methods.
\end{abstract}

\section{Introduction}

\label{sec:intro}

Recently, Generative Spoken Language Modeling (GSLM; \cite{lakhotiaGenerativeSpokenLanguage2021}) has emerged as a new paradigm for spoken language processing.
GSLM only requires speech resource for training, enabling language modeling without textual transcription.
This is achieved by training LMs on top of discrete representations extracted from raw audio (hereafter referred to as ``discrete units'').
As a result, GSLM broadens the scope of speech modeling to scenarios such as unwritten languages and infant language acquisition, where learning begins from auditory input before literacy~\cite{lavechinBabySLMLanguageacquisitionfriendlyBenchmark2023,lavechinSimulatingEarlyPhonetic2025}.
When combined with textual modalities, GSLM also serves as a foundation for cutting-edge speech models, including text-to-speech systems~\cite{VALL-E,zhang2024speechgpt} and spoken dialogue models~\cite{kyutai2024moshi,fang-etal-2025-llama}.
However, when trained solely on speech, its performance drops substantially in both language understanding and generation~\cite{hassidTextuallyPretrainedSpeech2023,parkLongFormSpeechGeneration2025,lin-etal-2025-align}.
Crucially, speech-only GSLMs scale more slowly than textual LMs~\cite{cuervoScalingPropertiesSpeech2024}, highlighting the need for more efficient training methods.

To address this issue, one possible approach is decreasing the sequence length.
Conversion from a speech signal into a discrete unit sequence (s2u) is typically achieved by quantizing continuous representations from self-supervised learning (SSL) models~\cite{oordRepresentationLearningContrastive2019,baevskiWav2vec20Framework2020,hsuHuBERTSelfSupervisedSpeech2021}.
This results in sequences that are much longer than text, which severely affects the training of Transformer-based LMs as computation cost increases quadratically with respect to the sequence length~\cite{vaswaniAttention2017}.
Previous research has proposed methods to decrease sequence length~\cite{algayresGenerativeSpokenLanguage2023,baade2025syllablelm,cho2025sylber,dekelExploringBenefitsTokenization2024,visser25_interspeech,sanders25_interspeech}, most of which involves training the s2u model.
Kando et al.~\cite{kando25} explored the spoken language understanding capability of GSLMs under the training-free simple s2u methods at scale, suggesting that the performance is enhanced by decreasing the sequence length and increasing the cluster size simultaneously.

\begin{figure}[t]
\centering
\includegraphics[width=.38\textwidth]{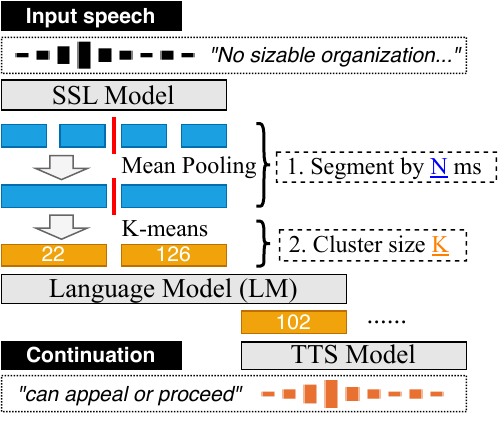}
\caption{
Overview of GSLM-based speech continuation.
It is achieved by synthesizing speech from discrete units predicted by the LM.
We varied segmentation width $N$ and cluster size $K$ at s2u step~\cite{kando25}.
Larger $N$ and smaller $K$ yield a lower bitrate.
}
\label{fig:overview}
\end{figure}

This paper investigates whether this observation also holds for GSLM-based speech synthesis.
Our experiments follow the same configurations as~\cite{kando25}.
GSLM-based speech synthesis consists of two tasks: speech resynthesis and speech continuation.
The task of speech resynthesis is to reproduce a given speech signal from discrete units, whereas speech continuation is to generate semantically and acoustically coherent speech conditioned on an input.
Figure~\ref{fig:overview} illustrates the overview of our speech continuation framework.
We segment the SSL representation sequence into fixed-length units of $N$ ms.
During quantization, we apply K-means clustering with varying numbers of clusters $K$.

Our comparative experiments show that while smaller segmentation width $N$ is generally preferred for speech resynthesis, moderately large $N$ values (i.e., lower bitrate settings) do not substantially degrade quality.
For speech continuation, several settings with moderately large $N$ achieve comparable performance to the baseline at lower bitrate.
These findings are supported by multiple evaluation metrics, including conventional metrics, LLM-based pairwise judgments, and human evaluations.
Correlation analysis with large-scale human subjective scores further shows that LLM-based metrics correlate better than conventional metrics.
However, the overall correlations remain limited, highlighting the need for more robust evaluation methods for speech continuation.

\section{Background: GSLM and its Evaluation}

GSLM consists of the three components:  
\textbf{(1) speech2unit (s2u)} converts speech into discrete units;  
\textbf{(2) unitLM (uLM)} is trained on these discrete units; and  
\textbf{(3) unit2speech (u2s)} synthesizes speech from the discrete units.  
In this paper, we examine the performance of u2s that involves two tasks: speech resynthesis and speech continuation.

\subsection{Task 1. Speech Resynthesis}

The task of speech resynthesis involves first applying s2u to obtain the unit sequence $\bm{u}$ from the given speech $\bm{s}$, and then applying u2s to synthesize speech $\bm{s'}$ from $\bm{u}$. 
In many studies (including the original GSLM paper), the quality of speech resynthesis is assessed using two metrics: \textbf{WER}\footnote{
CER is used in the original GSLM paper~\cite{lakhotiaGenerativeSpokenLanguage2021}, but we use WER instead as we conduct experiments on English throughout the paper.
}
and \textbf{UTMOS}~\cite{saeki22c_interspeech}.
\textbf{WER} measures \textit{intelligibility}, whether the linguistic content is preserved through the s2u and u2s pipeline. 
We calculate the WER between the gold transcription of $\bm{s}$ and the Automatic Speech Recognition (ASR) output of $\bm{s'}$. 
\textbf{UTMOS} is a prediction model of human Mean Opinion Score (MOS), which assesses the naturalness of speech audio.
In this study, we report \textbf{MCD} and \textbf{LogF0 RMSE} as complementary metrics to evaluate speech feature reconstruction.

\subsection{Task 2. Speech Continuation}

The conventional metrics for assessing speech continuation are \textbf{Perplexity (PPL)} and \textbf{diVERciTy (VERT)}~\cite{lakhotiaGenerativeSpokenLanguage2021}. 
\textbf{PPL} measures the semantic coherency of the generated contents, while \textbf{VERT} measures their diversity. 
Given a speech continuation dataset $S=\{\bm{s}_i\}_{i=1}^n$, we obtain a corresponding transcription dataset $T=\{\bm{t}_i\}_{i=1}^n$ using an ASR system, which is then used to compute both metrics.
\textbf{PPL} of $T$ is calculated by inputting $\bm{t}_i$ into an arbitrary textual LM.
Since a well-trained LM assigns low perplexity to fluent text, PPL serves as a metric for the meaningfulness or coherence of the generated content.  
\textbf{VERT} is defined as the geometric mean of self-BLEU and auto-BLEU~\cite{lakhotiaGenerativeSpokenLanguage2021}.
self-BLEU quantifies the overlaps between generated texts, indicating the diversity of them.
At low temperatures, uLM tends to generate repetitions such as ``fin, fin, fin,'' but self-BLEU alone cannot penalize such cases.  
auto-BLEU addresses this by quantifying the ratio of repeated $n$-grams within each sentence.

As pointed out by Lakhotia et al.~\cite{lakhotiaGenerativeSpokenLanguage2021}, there is a trade-off between PPL and VERT with respect to the temperature of uLM. 
This trend is illustrated in Figure~\ref{fig:tradeoff}. 
At low temperatures, uLM tends to generate uniform content or repetitions, resulting in lower PPL and higher VERT. 
In contrast, at high temperatures, uLM tends to generate random content, leading to higher PPL and lower VERT. 
Therefore, it is necessary to tune across multiple temperature values.

While PPL and VERT are primary used for assessing the quality of speech continuation, previous study points out that they are highly sensitive to many factors~\cite{hassidTextuallyPretrainedSpeech2023}.
Hence, more sophisticated evaluations such as human evaluation (\textbf{MMOS}; Meaningfulness MOS) or \textbf{pairwise evaluation with LLM-as-a-Judge}~\cite{parkLongFormSpeechGeneration2025} are proposed.
However, there is no consensus on which automatic metrics align with human subjective evalution.

\section{Background: Varying Segmentation Width and Cluster Size in s2u}

Kando et al.~\cite{kando25} explored the spoken language understanding capability of GSLM under different s2u methods.
As illustrated in Figure~\ref{fig:overview}, discrete units are obtained by segmenting continuous speech representations by $N$ ms, mean-pooling them, and then applying K-means clustering with cluster size $K$.
Afterward, units are deduplicated (e.g., 54 54 54 88 88 3 $\rightarrow$ 54 88 3). 
Spoken language understanding capability is assessed in a zero-shot setting: given a pair of utterances, the model is tested on whether it assigns higher likelihood to the correct one.
Experimental results generally suggest a positive effect of moderately large $N$ and larger $K$.
Moreover, these configurations reduce both training data size and runtime, as larger $N$ leads to a shorter sequence.

\section{Experimental Setting}

Figure~\ref{fig:overview} displays the overview of our experiment.
Following~\cite{kando25}, we obtain discrete units with various bitrates at the s2u step.
We investigate the impact of different s2u configurations on speech resynthesis and continuation.

\begin{figure}[t]
\centering
\begin{minipage}[b]{0.49\columnwidth}
    \centering
    \includegraphics[width=\columnwidth]{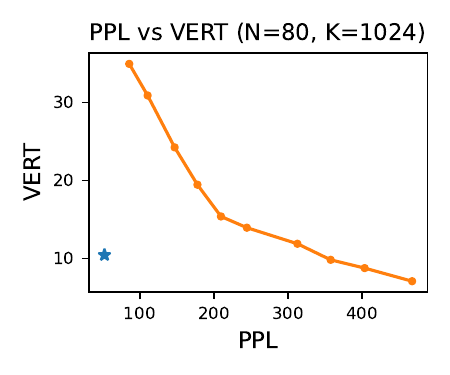}
\end{minipage}
\begin{minipage}[b]{0.49\columnwidth}
    \centering
    \includegraphics[width=\columnwidth]{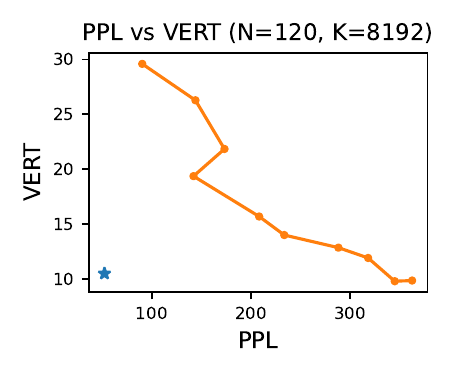}
\end{minipage}
\caption{
Trade-off between PPL and VERT with respect to temperature. 
Moving to the right indicates higher temperature. 
The blue star denotes the oracle value. 
A clear trade-off is observed in the left panel, but not in the right panel, which makes it difficult to calculate the area under the curve. 
}
\label{fig:tradeoff}
\end{figure}

\subsection{Training}

For s2u, we employ HuBERT-base~\cite{hsuHuBERTSelfSupervisedSpeech2021} as an SSL model and extracted representations from the 9th layer following~\cite{cho2025sylber}.
Since the frame shift width of HuBERT is 20 ms, $N$ is set to multiples of 20.
We experiment with eight values: \{20, 40, 80, 120, 160, 200, 240, 280\}, where $N=20$ corresponds to the original HuBERT representation. 
Since $N=20$ settings are commonly employed in GSLM-based model family~\cite{lakhotiaGenerativeSpokenLanguage2021,hassidTextuallyPretrainedSpeech2023,kharitonov-etal-2022-text}, we treat this as the baseline.
For quantization, we trained K-means models on the 100-hour clean subset of LibriSpeech~\cite{vassilLibrispeech2015} using eight $K$ values of powers of two: from $2^7=128$ to $2^{14}=16384$.
In total, we employ $8\times8=64$ s2u methods.
For uLM, we trained OPT~\cite{zhangOPT2022}, a Transformer-based autoregressive LM, with the same parameter setings as~\cite{kando25}.
uLM is trained on the full LibriSpeech training set (960 hours).
For u2s, we can apply any TTS models in principle.
We hypothesize that the training efficiency of a TTS model depends on the discrete units (of which there are 64 variations). 
Therefore, we employed two TTS models for comparison. 
The first is Tacotron2~\cite{tacotron2} with a Parallel WaveGAN vocoder~\cite{parallel}. 
The second is VITS~\cite{vits}, an end-to-end model from units to speech. 
We trained u2s models on LJSpeech~\cite{ljspeech17}.
All u2s training was conducted using ESPNet~\cite{watanabe2018espnet}.

\subsection{Evaluation}

\begin{figure}[t]
\centering
\includegraphics[width=.47\textwidth]{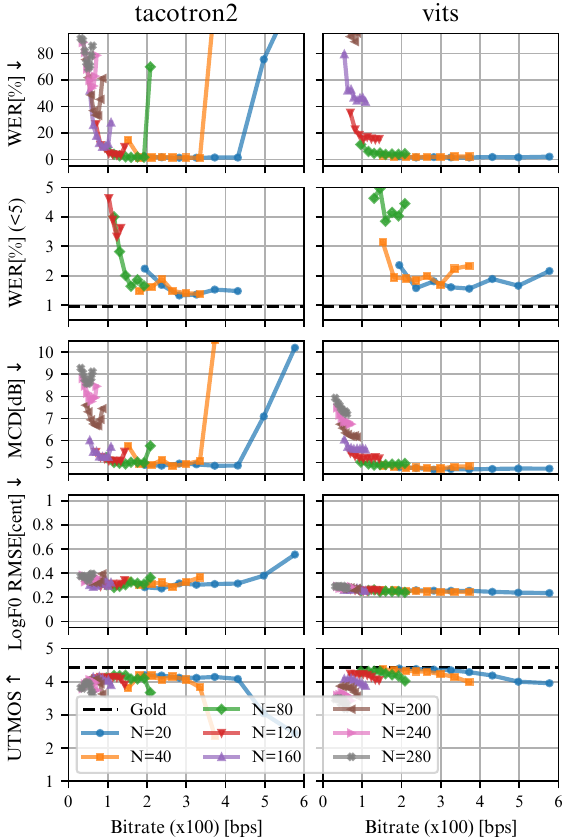}
\caption{
Results of speech resynthesis.
The x-axis represents the bitrate [bps] of each setting, calculated by multiplying unit entropy by the average number of units per second~\cite{dunbar19_interspeech}.
Moving to the right indicates higher $K$.
The second row presents a magnified view of the results with WER below 5\%. 
}
\label{fig:resynthesis_score}
\end{figure}

For all evaluations, we transcribed resynthesized and continued speech using \texttt{openai/whisper-large-v3}.
PPL values were calculated with \texttt{meta-llama/Llama-3.1-8B}.
For speech continuation, we prepared input speech from the LJSpeech development set. 
We extracted utterances longer than six seconds and fed the speech units corresponding to the first three seconds into the uLM. 
We continued generation for seven seconds, resulting in 10 seconds audio.
We calculated PPL $P_t$ and VERT $V_t$ for ten temperature values $t\in\{0.3, 0.4, \ldots, 1.2\}$.
PPL and VERT values of gold transcriptions of LJSpeech validation set are regarded as oracles ($P_o$ and $V_o$, respectively).
In the original GSLM paper~\cite{lakhotiaGenerativeSpokenLanguage2021}, the score for speech continuation is quantified by the area under curve of $(P_t, V_t)$ plots.
Although it presupposes that $P_t$ and $V_t$ change monotonically, we observed that several settings violate this condition (Figure~\ref{fig:tradeoff}).
Therefore, we instead applied min-max normalization and calculated the Euclidean distance between $(P_t, V_t)$ and $(P_o, V_o)$. 
The temperature with the shortest distance was regarded as optimal for the corresponding $(N, K)$ setting.
Experimental codes are publicly available\footnote{\url{https://github.com/gifdog97/espnet/tree/master/egs2/ljspeech/tts1/myscripts}}.

\section{Results}

\subsection{Speech Resynthesis}
\label{ssec:result-resynthesis}

Figure~\ref{fig:resynthesis_score} presents the speech resynthesis results.
Overall, although larger $N$ degrades performance, moderately large values such as 40 or 80 achieve performance comparable to the baseline setting ($N=20$).
This indicates that speech resynthesis quality can be preserved at lower bitrates.
In terms of the selection of u2s models, we observe clear differences between Tacotron2 and VITS.
In terms of intelligibility (WER), Tacotron2 consistently outperforms VITS at the same $N$ values, except for outliers at $N=20$ and $40$.
This trend is generally reversed for other metrics (MCD, LogF0 RMSE, and UTMOS), suggesting that VITS achieves better acoustic quality.
Overall, Tacotron2 and VITS demonstrate complementary strengths, with Tacotron2 excelling in intelligibility and VITS in acoustic fidelity.
Interestingly, particularly for VITS, LogF0 RMSE remains low even when other metrics deteriorate.
This behavior may stem from the monotonic alignment search mechanism in VITS, which promotes stable temporal alignment and could help maintain pitch contour even when overall acoustic fidelity declines.

\begin{figure}[t]
\centering
\includegraphics[width=.47\textwidth]{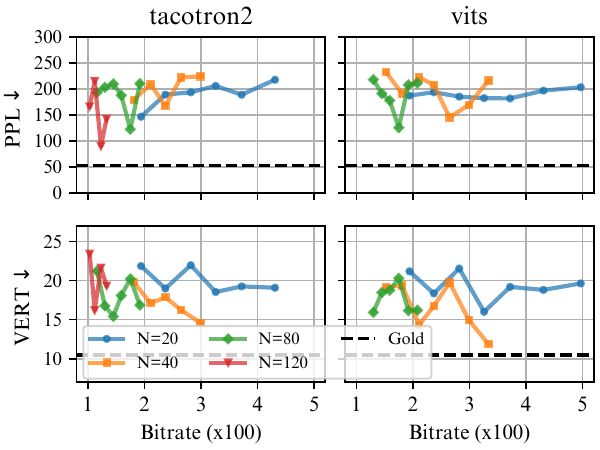}
\caption{
Results of speech continuation.
Only settings where speech resynthesis scores meet the conditions of WER below 5 and UTMOS above 4 are shown.
}
\label{fig:continuation_score}
\end{figure}


\subsection{Speech Continuation}
\label{ssec:result-continuation}

To ensure the quality of the synthesized speech, we evaluated speech continuation only under settings with WER below 5 and UTMOS above 4. 

\subsubsection{PPL and VERT}

The results are shown in Figure~\ref{fig:continuation_score}. 
For both metrics, settings with larger $N$ perform on par with, or slightly better than, the baseline.
We also observe a discrepancy between the results of speech resynthesis and speech continuation. 
For speech resynthesis, the baseline setting ($N=20$) generally yields the best performance, whereas this is not the case for speech continuation. 
On the other hand, several settings that are on par with the baseline in speech continuation (e.g., $(120, 2^{13})$ in Tacotron2, corresponding to the rightmost point) exhibit worse WER values in resynthesis.
This discrepancy suggests that the optimal unit configuration is task-dependent, highlighting the need for speech representations that balance phonetic fidelity with semantic modeling.
Interestingly, we observe fluctuations with respect to $K$, which may be an artifact of the min-max normalization.

\subsubsection{LLM-based Evaluation}

\begin{figure}[t]
\centering
\includegraphics[width=.47\textwidth]{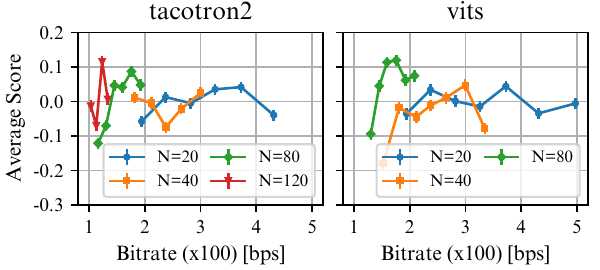}
\caption{
Average scores of LLM-based pairwise evaluation.
}
\label{fig:pairwise_score}
\end{figure}

As a more sophisticated evaluation, we conducted a pairwise evaluation using the LLM-as-a-Judge framework~\cite{zhengLLM2023}.
Referring to \cite{parkLongFormSpeechGeneration2025}, we designed a prompt to judge which of two continuation texts (A or B) is superior, using \textit{fluency}, \textit{coherence}, and \textit{logicality} as guiding criteria for the overall decision. 
The judgment score is defined on a five-point scale: 
$1$ if A is significantly better than B, 
$0.5$ if A is slightly better,
$0$ if they are tied,
$-0.5$ if B is slightly better,
and $-1$ if B is significantly better.
We tested on all the possible pairs of speech continuation settings and averaged the scores.
We used \texttt{GPT-4.1-mini}\footnote{\url{https://openai.com/index/gpt-4-1/}} as an LLM.
Figure~\ref{fig:pairwise_score} shows average scores of each setting.
Both Tacotron2 and VITS results indicate that settings with moderately large $N$ (80–120) and large $K$ outperform the baseline setting ($N=20$). 
Notably, the best performance is achieved at lower bitrate ($(120,4096)$ in Tacotron2, $(80, 4096)$ in VITS).


\begin{figure}[t]
\centering
\includegraphics[width=.47\textwidth]{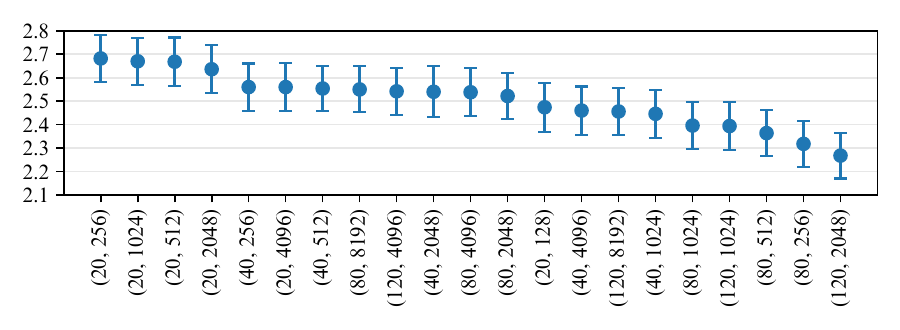}
\caption{
MMOS with 95\% CI.
}
\label{fig:mmos}
\end{figure}

\subsubsection{Influence of u2s Model on Speech Continuation}
\label{sssec:tts}

We further examine whether the choice of TTS model influences speech continuation quality.
To quantify model-dependent effects,
we synthesized speech audios using Tacotron2 and VITS from the set of identical discrete unit sequences,
then compute Spearman’s rank correlations of PPL and VERT between them\footnote{
We don't consider LLM-as-a-Judge as it involves comparison beween different settings that introduces other factor.
}.
The resulting correlation coefficients are 0.760 (PPL) and 0.804 (VERT) at a level of $p<0.05$.
This suggests that the quality of the synthesized speech content is largely preserved across TTS models, indicating limited impact of the choice of models on linguistic content.

\subsubsection{Human Evaluation}

We conducted human evaluation on speech continuation using Meaningfulness MOS (MMOS; \cite{lakhotiaGenerativeSpokenLanguage2021}).
We ask human raters to rate how meaningful a given audio sample is on a 1–5 scale with increments of 1.
We choose all the 21 settings from Tacotron2, extract 50 samples per setting, and enforce 10 raters per sample.
Figure~\ref{fig:mmos} shows the result.
Although baseline settings with $N=20$ are superior to others, there is no significant differences between settings each other from $(20, 256)$ to $(40, 1024)$\footnote{
Significance tests are conducted using Mann-Whitney U test~\cite{rosenberg17_interspeech} at $p<0.05$ with Bonferroni correction for multiple comparisons.
}.

To identify the best setting, we additionally conducted an AB test across different $N$ values.
We selected $(20, 256)$, $(40, 256)$, $(80, 4096)$, and $(120, 4096)$, which performed best in MMOS or LLM-as-a-Judge.
Human raters are presented a pair of speech audio from different two settings and judge which is more meaningful.
We extract 20 samples with the same prompt and enforce 6 raters per pair.
Result in Table~\ref{tab:abtest} shows that $(20, 256)$ and $(80, 4096)$ perform better with statistical significance.
We conclude that speech continuation quality can be maintained at larger $N$ settings with lower bitrate.

\subsubsection{Correlation between Subjective and Automatic Metrics}

To assess the quality of automatic evaluation metrics (PPL, VERT, and LLM-as-a-Judge) for speech continuation, we calculated the correlation between MMOS and them.
Table~\ref{tab:srcc_mmos} shows the result.
While LLM-as-a-Judge shows the highest correlation, the value is still low.
This highlights the need for more robust automatic evaluation method for evaluating speech meaningfulness.

\input{table/abtest}
\input{table/srcc_mmos}


%

\section{Conclusion}
\label{sec:conclusion}

In this paper, we investigated GSLM-based speech resynthesis and continuation under discrete representations with varying bitrates.
By controlling segmentation width and K-means cluster size, we systematically analyzed the impact of bitrate on synthesis quality.
Our results show that both intelligible speech resynthesis and high-quality continuation can be achieved at substantially lower bitrates than the conventional setting, suggesting that the commonly used bitrate may contain redundant information for speech generation.
Furthermore, although LLM-based metrics correlate better with human judgments than conventional metrics, the correlation remains limited, highlighting the need for more reliable automatic evaluation methods.

\clearpage

\section{Acknowledgments}
This work was supported by
JST ACT-X Grant Number JPMJAX24C9 and
JSPS KAKENHI Grant Number 26KJ0792
.

\section{Generative AI use disclosure}
ChatGPT was used for language polishing.

\bibliographystyle{IEEEtran}
\bibliography{mybib}

\end{document}

%% file: table/abtest.tex
\begin{table}[t]
\centering
\caption{
Win rate of AB test by human evaluator.
* indicates $p<0.05$ under binomial test.
}
\label{tab:abtest}
\begin{tabular}{lllll}
\toprule
           & $(40, 256)$ & $(80, 4096)$ & $(120, 4096)$ \\
\midrule
$(20, 256)$  &  0.500 & 0.558 & 0.617* \\
$(40, 256)$  &        & 0.483 & 0.575 \\
$(80, 4096)$ &        &       & 0.671* \\
\bottomrule
\end{tabular}
\end{table}

%% file: table/srcc_mmos.tex
\begin{table}[t]
\centering
\caption{Spearman rank correlation coefficients (SRCC) between MMOS and automatic metrics. * indicates $p<0.05$.}
\label{tab:srcc_mmos}
\begin{tabular}{lcccc}
\toprule
        & PPL & VERT & LLM  \\
\midrule
MMOS    & $-0.105$* & 0.031 & \textbf{0.323}*  \\
\bottomrule
\end{tabular}
\end{table}

%% file: mybib.bib
@article{lakhotiaGenerativeSpokenLanguage2021,
  title = {On {{Generative Spoken Language Modeling}} from {{Raw Audio}}},
  author = {Lakhotia, Kushal and Kharitonov, Eugene and Hsu, Wei-Ning and Adi, Yossi and Polyak, Adam and Bolte, Benjamin and Nguyen, Tu-Anh and Copet, Jade and Baevski, Alexei and Mohamed, Abdelrahman and Dupoux, Emmanuel},
  editor = {Roark, Brian and Nenkova, Ani},
  year = {2021},
  journal = {TACL},
  publisher = {MIT Press},
  address = {Cambridge, MA},
  doi = {10.1162/tacl_a_00430},
  urldate = {2023-12-18},
  volume = "9",
  pages = "1336--1354"
}

@article{lavechinSimulatingEarlyPhonetic2025,
  title = {Simulating {{Early Phonetic}} and {{Word Learning Without Linguistic Categories}}},
  author = {Lavechin, Marvin and {de Seyssel}, Maureen and Titeux, Hadrien and Wisniewski, Guillaume and Bredin, Herv{\'e} and Cristia, Alejandrina and Dupoux, Emmanuel},
  year = {2025},
  journal = {Developmental Science},
  Volume = {28},
  Number = {2},
  issn = {1467-7687},
  doi = {10.1111/desc.13606},
  urldate = {2025-02-13},
  copyright = {{\copyright} 2025 John Wiley \& Sons Ltd.},
  langid = {english}
}

@inproceedings{lavechinBabySLMLanguageacquisitionfriendlyBenchmark2023,
  title = {{{BabySLM}}: Language-Acquisition-Friendly Benchmark of Self-Supervised Spoken Language Models},
  shorttitle = {{{BabySLM}}},
  booktitle = {{Interspeech}},
  author = {Lavechin, Marvin and Sy, Yaya and Titeux, Hadrien and Bland{\'o}n, Mar{\'i}a Andrea Cruz and R{\"a}s{\"a}nen, Okko and Bredin, Herv{\'e} and Dupoux, Emmanuel and Cristia, Alejandrina},
  year = {2023},
  doi = {10.21437/Interspeech.2023-978},
  urldate = {2024-03-20},
  langid = {english},
  pages     = {4588--4592}
}

@inproceedings{lin-etal-2025-align,
    title = "Align-{SLM}: Textless Spoken Language Models with Reinforcement Learning from {AI} Feedback",
    author = "Lin, Guan-Ting  and
      Shivakumar, Prashanth Gurunath  and
      Gourav, Aditya  and
      Gu, Yile  and
      Gandhe, Ankur  and
      Lee, Hung-yi  and
      Bulyko, Ivan",
    booktitle = "ACL (main)",
    year = "2025",
    doi = "10.18653/v1/2025.acl-long.997",
    ISBN = "979-8-89176-251-0",
    pages = "20395--20411"
}

@inproceedings{cuervoScalingPropertiesSpeech2024,
  title = {Scaling {{Properties}} of {{Speech Language Models}}},
  booktitle = {EMNLP (main)},
  author = {Cuervo, Santiago and Marxer, Ricard},
  year = {2024},
  doi = {10.18653/v1/2024.emnlp-main.21},
  urldate = {2024-12-18},
  pages = "351--361"
}

@inproceedings{hassidTextuallyPretrainedSpeech2023,
  title = {Textually {{Pretrained Speech Language Models}}},
  booktitle = {{{NeurIPS}}},
  author = {Hassid, Michael and Remez, Tal and Nguyen, Tu Anh and Gat, Itai and Conneau, Alexis and Kreuk, Felix and Copet, Jade and Defossez, Alexandre and Synnaeve, Gabriel and Dupoux, Emmanuel and Schwartz, Roy and Adi, Yossi},
  year = {2023},
  eprint = {2305.13009},
  primaryclass = {cs, eess},
  doi = {10.48550/arXiv.2305.13009},
  urldate = {2024-09-18},
  pages = {63483--63501},
  archiveprefix = {arXiv}
}

@inproceedings{kharitonov-etal-2022-text,
    title = "Text-Free Prosody-Aware Generative Spoken Language Modeling",
    author = "Kharitonov, Eugene  and
      Lee, Ann  and
      Polyak, Adam  and
      Adi, Yossi  and
      Copet, Jade  and
      Lakhotia, Kushal  and
      Nguyen, Tu Anh  and
      Riviere, Morgane  and
      Mohamed, Abdelrahman  and
      Dupoux, Emmanuel  and
      Hsu, Wei-Ning",
    booktitle = "ACL (main)",
    month = may,
    year = "2022",
    pages = "8666--8681",
    doi = "10.18653/v1/2022.acl-long.593"
}

@inproceedings{kando25,
  title     = {{Exploring the Effect of Segmentation and Vocabulary Size on Speech Tokenization for Speech Language Models}},
  author    = {Shunsuke Kando and Yusuke Miyao and Shinnosuke Takamichi},
  year      = {2025},
  booktitle = {{Interspeech}},
  doi       = {10.21437/Interspeech.2025-310},
  pages     = {5728-5732},
  issn      = {2958-1796},
}

@inproceedings{parkLongFormSpeechGeneration2025,
  title = {Long-{{Form Speech Generation}} with {{Spoken Language Models}}},
  booktitle = {{{ICML}}},
  author = {Park, Se Jin and Salazar, Julian and Jansen, Aren and Kinoshita, Keisuke and Ro, Yong Man and {Skerry-Ryan}, R. J.},
  year = {2025},
  urldate = {2025-07-16},
  langid = {english}
}

@inproceedings{dekelExploringBenefitsTokenization2024,
  title = {Exploring the {{Benefits}} of {{Tokenization}} of {{Discrete Acoustic Units}}},
  booktitle = {Interspeech},
  author = {Dekel, Avihu and Fernandez, Raul},
  year = {2024},
  pages = {2780--2784},
  eprint = {2406.05547},
  primaryclass = {cs, eess},
  urldate = {2024-06-11},
  archiveprefix = {arXiv},
  langid = {english}
}

@article{hsuHuBERTSelfSupervisedSpeech2021,
  title = {{{HuBERT}}: {{Self-Supervised Speech Representation Learning}} by {{Masked Prediction}} of {{Hidden Units}}},
  shorttitle = {{{HuBERT}}},
  author = {Hsu, Wei-Ning and Bolte, Benjamin and Tsai, Yao-Hung Hubert and Lakhotia, Kushal and Salakhutdinov, Ruslan and Mohamed, Abdelrahman},
  year = {2021},
  journal = {IEEE/ACM TASLP},
  page = {3451 -- 3460},
  volume = {29},
  issn = {2329-9290},
  doi = {10.1109/TASLP.2021.3122291},
  urldate = {2024-02-08}
}

@inproceedings{baevskiWav2vec20Framework2020,
  title = {Wav2vec 2.0: {{A Framework}} for {{Self-Supervised Learning}} of {{Speech Representations}}},
  shorttitle = {Wav2vec 2.0},
  booktitle = {NeurIPS},
  author = {Baevski, Alexei and Zhou, Yuhao and Mohamed, Abdelrahman and Auli, Michael},
  year = {2020},
  urldate = {2024-02-22},
  pages = {12449 -- 12460}
}

@article{oordRepresentationLearningContrastive2019,
  title = {Representation {{Learning}} with {{Contrastive Predictive Coding}}},
  author = {van den Oord, Aaron and Li, Yazhe and Vinyals, Oriol},
  year = {2019},
  journal = {arXiv preprint arXiv:1807.03748},
  doi = {10.48550/arXiv.1807.03748},
  urldate = {2025-02-07},
  archiveprefix = {arXiv}
}

@inproceedings{vaswaniAttention2017,
  author       = {Ashish Vaswani and
                  Noam Shazeer and
                  Niki Parmar and
                  Jakob Uszkoreit and
                  Llion Jones and
                  Aidan N. Gomez and
                  Lukasz Kaiser and
                  Illia Polosukhin},
  title        = {Attention is All you Need},
  booktitle    = {NeurIPS},
  year         = {2017},
  pages        = {6000 -- 6010},
  bibsource    = {dblp computer science bibliography, https://dblp.org}
}

@inproceedings{algayresGenerativeSpokenLanguage2023,
  title = {Generative {{Spoken Language Model}} Based on Continuous Word-Sized Audio Tokens},
  booktitle = {EMNLP (main)},
  author = {Algayres, Robin and Adi, Yossi and Nguyen, Tu and Copet, Jade and Synnaeve, Gabriel and Sagot, Beno{\^i}t and Dupoux, Emmanuel},
  year = {2023},
  doi = {10.18653/v1/2023.emnlp-main.182},
  pages = {3008--3028},
  urldate = {2024-05-16}
}

@inproceedings{
baade2025syllablelm,
title={Syllable{LM}: Learning Coarse Semantic Units for Speech Language Models},
author={Alan Baade and Puyuan Peng and David Harwath},
booktitle={ICLR},
year={2025},
}

@inproceedings{
cho2025sylber,
title={Sylber: Syllabic Embedding Representation of Speech from Raw Audio},
author={Cheol Jun Cho and Nicholas Lee and Akshat Gupta and Dhruv Agarwal and Ethan Chen and Alan Black and Gopala Anumanchipalli},
booktitle={ICLR},
year={2025},
}

@inproceedings{saeki22c_interspeech,
  title     = {{UTMOS: UTokyo-SaruLab System for VoiceMOS Challenge 2022}},
  author    = {{Takaaki Saeki and Detai Xin and Wataru Nakata and Tomoki Koriyama and Shinnosuke Takamichi and Hiroshi Saruwatari}},
  year      = {{2022}},
  booktitle = {{Interspeech}},
  doi       = {{10.21437/Interspeech.2022-439}},
  pages     = {4521--4525},
  issn      = {{2958-1796}},
}

@INPROCEEDINGS{tacotron2,
  author={Shen, Jonathan and Pang, Ruoming and Weiss, Ron J. and Schuster, Mike and Jaitly, Navdeep and Yang, Zongheng and Chen, Zhifeng and Zhang, Yu and Wang, Yuxuan and Skerrv-Ryan, Rj and Saurous, Rif A. and Agiomvrgiannakis, Yannis and Wu, Yonghui},
  booktitle={ICASSP}, 
  title={Natural TTS Synthesis by Conditioning Wavenet on MEL Spectrogram Predictions}, 
  year={2018},
  pages={4779--4783},
  volume={},
  number={},
  doi={10.1109/ICASSP.2018.8461368}
}

@InProceedings{vits,
  title = 	 {Conditional Variational Autoencoder with Adversarial Learning for End-to-End Text-to-Speech},
  author =       {Kim, Jaehyeon and Kong, Jungil and Son, Juhee},
  booktitle = 	 {ICML},
  year = 	 {2021}
}

@INPROCEEDINGS{parallel,
  author={Yamamoto, Ryuichi and Song, Eunwoo and Kim, Jae-Min},
  booktitle={ICASSP}, 
  title={Parallel Wavegan: A Fast Waveform Generation Model Based on Generative Adversarial Networks with Multi-Resolution Spectrogram}, 
  year={2020},
  pages={6199-6203},
  volume={},
  number={},
  doi={10.1109/ICASSP40776.2020.9053795}
}

@misc{ljspeech17,
  author       = {Keith Ito and Linda Johnson},
  title        = {The LJ Speech Dataset},
  howpublished = {\url{https://keithito.com/LJ-Speech-Dataset/}},
  year         = 2017
}

@inproceedings{watanabe2018espnet,
  author={Shinji Watanabe and Takaaki Hori and Shigeki Karita and Tomoki Hayashi and Jiro Nishitoba and Yuya Unno and Nelson {Enrique Yalta Soplin} and Jahn Heymann and Matthew Wiesner and Nanxin Chen and Adithya Renduchintala and Tsubasa Ochiai},
  title={{ESPnet}: End-to-End Speech Processing Toolkit},
  year={2018},
  booktitle={Interspeech},
  pages={2207--2211},
  doi={10.21437/Interspeech.2018-1456},
}

@inproceedings{vassilLibrispeech2015,
  author       = {Vassil Panayotov and
                  Guoguo Chen and
                  Daniel Povey and
                  Sanjeev Khudanpur},
  title        = {Librispeech: An {ASR} corpus based on public domain audio books},
  booktitle    = {ICASSP},
  year         = {2015},
  doi          = {10.1109/ICASSP.2015.7178964},
  pages        = {5206-5210},
  bibsource    = {dblp computer science bibliography, https://dblp.org}
}

@article{zhangOPT2022,
  author       = {Susan Zhang and
                  Stephen Roller and
                  Naman Goyal and
                  Mikel Artetxe and
                  Moya Chen and
                  Shuohui Chen and
                  Christopher Dewan and
                  Mona T. Diab and
                  Xian Li and
                  Xi Victoria Lin and
                  Todor Mihaylov and
                  Myle Ott and
                  Sam Shleifer and
                  Kurt Shuster and
                  Daniel Simig and
                  Punit Singh Koura and
                  Anjali Sridhar and
                  Tianlu Wang and
                  Luke Zettlemoyer},
  title        = {{OPT:} Open Pre-trained Transformer Language Models},
  year         = {2022},
  doi          = {10.48550/ARXIV.2205.01068},
  journal      = {arXiv preprint: arXiv:2205.01068},
  bibsource    = {dblp computer science bibliography, https://dblp.org}
}

@inproceedings{
zhengLLM2023,
title={Judging {LLM}-as-a-Judge with {MT}-Bench and Chatbot Arena},
author={Lianmin Zheng and Wei-Lin Chiang and Ying Sheng and Siyuan Zhuang and Zhanghao Wu and Yonghao Zhuang and Zi Lin and Zhuohan Li and Dacheng Li and Eric Xing and Hao Zhang and Joseph E. Gonzalez and Ion Stoica},
booktitle={NeurIPS Datasets and Benchmarks Track},
year={2023},
}

@inproceedings{visser25_interspeech,
  title     = {{Spoken Language Modeling with Duration-Penalized Self-Supervised Units}},
  author    = {Nicol Visser and Herman Kamper},
  year      = {2025},
  booktitle = {{Interspeech}},
  doi       = {10.21437/Interspeech.2025-340},
  pages     = {1968-1972},
  issn      = {2958-1796},
}

@inproceedings{sanders25_interspeech,
  title     = {{Segmentation-Variant Codebooks for Preservation of Paralinguistic and Prosodic Information}},
  author    = {Nicholas Sanders and Yuanchao Li and Korin Richmond and Simon King},
  year      = {2025},
  booktitle = {{Interspeech}},
  doi       = {10.21437/Interspeech.2025-2075},
  pages     = {5403-5407},
  issn      = {2958-1796},
}

@inproceedings{rosenberg17_interspeech,
  title     = {{Bias and Statistical Significance in Evaluating Speech Synthesis with Mean Opinion Scores}},
  author    = {Andrew Rosenberg and Bhuvana Ramabhadran},
  year      = {2017},
  booktitle = {{Interspeech}},
  pages     = {3976--3980},
  doi       = {10.21437/Interspeech.2017-479},
  issn      = {2958-1796},
}

@inproceedings{dunbar19_interspeech,
  title     = {{The Zero Resource Speech Challenge 2019: TTS Without T}},
  author    = {Ewan Dunbar and Robin Algayres and Julien Karadayi and Mathieu Bernard and Juan Benjumea and Xuan-Nga Cao and Lucie Miskic and Charlotte Dugrain and Lucas Ondel and Alan W. Black and Laurent Besacier and Sakriani Sakti and Emmanuel Dupoux},
  year      = {2019},
  booktitle = {{Interspeech}},
  pages     = {1088--1092},
  doi       = {10.21437/Interspeech.2019-2904},
  issn      = {2958-1796},
}

@techreport{kyutai2024moshi,
      title={Moshi: a speech-text foundation model for real-time dialogue},
      author={Alexandre D\'efossez and Laurent Mazar\'e and Manu Orsini and
      Am\'elie Royer and Patrick P\'erez and Herv\'e J\'egou and Edouard Grave and Neil Zeghidour},
      year={2024},
      eprint={2410.00037},
      archivePrefix={arXiv},
      primaryClass={eess.AS},
}

@inproceedings{fang-etal-2025-llama,
    title = "{LL}a{MA}-Omni 2: {LLM}-based Real-time Spoken Chatbot with Autoregressive Streaming Speech Synthesis",
    author = "Fang, Qingkai  and
      Zhou, Yan  and
      Guo, Shoutao  and
      Zhang, Shaolei  and
      Feng, Yang",
    booktitle = "ACL (long)",
    year = "2025",
    pages = "18617--18629",
    doi = "10.18653/v1/2025.acl-long.912",
    ISBN = "979-8-89176-251-0"
}

@article{zhang2024speechgpt,
    title={SpeechGPT-Gen: Scaling Chain-of-Information Speech Generation}, 
    author={Dong Zhang and Xin Zhang and Jun Zhan and Shimin Li and Yaqian Zhou and Xipeng Qiu},
    year={2024},
    journal = {arXiv preprint arXiv:2401.13527},
}

@ARTICLE{VALL-E,
  author={Chen, Sanyuan and Wang, Chengyi and Wu, Yu and Zhang, Ziqiang and Zhou, Long and Liu, Shujie and Chen, Zhuo and Liu, Yanqing and Wang, Huaming and Li, Jinyu and He, Lei and Zhao, Sheng and Wei, Furu},
  journal={IEEE Transactions on Audio, Speech and Language Processing}, 
  title={Neural Codec Language Models are Zero-Shot Text to Speech Synthesizers}, 
  year={2025},
  volume={33},
  number={},
  pages={705-718},
  doi={10.1109/TASLPRO.2025.3530270}
}
